RESEARCH ARTICLE

# Artificial Intelligence Governance For Businesses

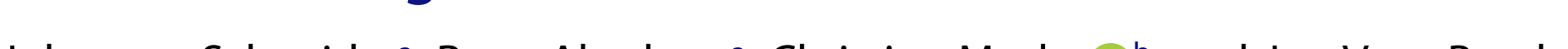

Johannes Schneider[a], Rene Abraham[a], Christian Meske[b], and Jan Vom Brocke[a]

[a]Institute of Information Systems, University of Liechtenstein, Vaduz, Liechtenstein; [b]Ruhr University of Bochum, Bochum, Germany

**ABSTRACT**
While artificial intelligence (AI) governance is thoroughly discussed on a philosophical, societal, and regulatory level, few works target companies. We address this gap by deriving a conceptual framework from literature. We decompose AI governance into governance of data, machine learning models, and AI systems along the dimensions of who, what, and how "is governed." This decomposition enables the evolution of existing governance structures. Novel, business-specific aspects include measuring data value and novel AI governance roles.

## Introduction

Artificial Intelligence (AI) has emerged as an essential field in research and practice, with projected spending of almost 100 billion US dollars by 2023 from 38 billion in 2019 (Shirer & Daquila, 2019). AI exhibits forms of intelligent behavior, allowing for an extensive range of cost-efficient, well-performing applications. While AI incorporates many techniques, in this work, we understand AI as systems that learn from samples, i.e., these systems rely on models stemming from a subbranch of AI denoted "machine learning." Machine learning (ML) includes techniques such as deep learning that infer decision-making behavior from data. It is responsible for most successes of AI since learning from data rather than extracting and implementing domain expert rules yields AI systems of superior performance at relatively low costs. For example, AI has shown remarkable success in many application areas (Meske et al., 2022), for instance, job recruitment (Pan et al., 2022), credit scoring (H. Wang et al., 2019), designing floorplans for microchips (Khang, 2021), managing predictive maintenance strategies (Arena et al., 2022), autopilots in aviation (Garlick, 2017), or autonomous driving (Grigorescu et al., 2020). Thus, AI is also the focus of many organizations, and research shows that 90% of executives agree that AI is a business opportunity critical to their company's success (Ransbotham et al., 2019). At the same time, only 10% of the executives report a significant financial gain from implementing AI (Ransbotham et al., 2020). Consequently, there is still a high level of uncertainty about how AI technology can be used effectively to generate value in organizations and how to specifically leverage the technology to make profits for organizations.

While AI is not new, its recent technical and regulatory advances are staggering (Burt, 2021), and they are likely to remain high-paced for some time. Fast advancements in AI make it difficult for companies to keep up and identify appropriate governance mechanisms to reap economic benefits of AI. At the same time, companies must comply with a growing body of regulations touching upon data, ML models and AI systems. Moreover, AI exhibits characteristics that make it both attractive but also challenging to rely on, for instance:

**(1) The output of AI is often difficult to understand** (Adadi & Berrada, 2018), even for applications where its models are easy to build (e.g., using AutoML or adapting existing models). It is hard to comprehend why AI makes a specific decision and how an AI system works in general. The lack of understanding poses challenges to meeting regulatory requirements and improving systems beyond what is already known.

**(2) AI produces unexpected results that are partly beyond the control of an organization**. It exhibits non-predictable, "ethics"-unaware, data-induced behavior yielding novel security, safety, and fairness issues. AI provides largely statistical guarantees on its performance. AI might malfunction in scenarios that are hard to foresee and exploitable by malicious agents, e.g., as demonstrated for existing self-driving cars (Morgulis et al., 2019). An AI model is often trained using objectives such as maximizing overall accuracy

**CONTACT** Johannes Schneider johannes.schneider@uni.li University of Liechtenstein, Vaduz, Liechtenstein,

and might be based on imbalanced data, e.g., image data that contains more samples of one race than another. This might lead to unethical models that exhibit, for example, a much higher error rate for one race than for another.

**(3) Data may lead to biases in decisions of AI systems**. Systems with biased decision-making pose a reputation risk as witnessed, e.g., for Amazon's AI-based recruitment tool (Dastin, 2018). This could ultimately lead to legal consequences. While such systems often cause public outcries, they are no surprise. ML models – forming the brain of AI systems – are typically optimized toward a single metric (Goodfellow et al., 2016). They commonly focus exclusively on performance, e.g., they maximize the number of correct decisions, ignoring other concerns such as the rationale and information on what these decisions should or should not be based on. This highlights that mechanisms (and governance) to ensure adequate decision-making are needed.

**(4) AI is quickly evolving on a technological level**, constantly yielding new business opportunities and possibilities for digital innovation. AI components have entered many products and industries that are traditionally unaware of AI technology. It also yields new standards (Cihon, 2019) and new regulatory requirements (Burt, 2021). While today AI does not exhibit "general intelligence," it is unknown when it could emerge (Ford, 2018). AI might also undermine human authority, which is one reason for likely future regulations, e.g., related to human agency and oversight (European Commission, 2020).

Hence, governance mechanisms play an important role to mitigate AI challenges and raise AI potential in organizations. Governance is strongly positively related to firm performance (Bhagat & Bolton, 2008). AI governance for business comprises of the structure of rules, practices, and processes used to ensure that the organization's AI technology sustains and extends the organization's strategies and objectives. Our goal is to provide a governance perspective for businesses on a tangible level that highlights relevant governance concepts and sets the boundaries and practices for the successful use of AI to meet a company's objectives, such as profitability and efficiency. At the same time, AI adoption and capabilities are still under research (e.g., Jöhnk et al., 2021; Mikalef & Gupta, 2021). The focus of this article is not on temporary AI governance during a transient adoption phase but on (long-term) governance of AI. Aside from regulatory aspects, there are also technological contributions to AI that impact governance. They allow governing in novel ways or ways that at least shed some light to what extent governance mechanisms are realistic to implement. In particular, we elaborate on model governance in this work, including testing, data valuation, and data quality issues.

While there is some debate on the disruptiveness of AI on governance (Liu & Maas, 2021), current governance structures should be considered both from a research and practitioner perspective. Thus, we approach AI governance from the lens of existing information technology (IT) and data governance frameworks. For instance, there exist models for AI governance, i.e., a layered model for AI governance (Gasser & Almeida, 2017), consisting of a social and legal layer (covering norms, regulation, and legislation), an ethical layer (criteria and principles), and a technical layer (data governance, accountability, standards), hence focusing on a strong societal and humanistic aspect. However, existing approaches neglect corporate and practical concerns. With this article, we aim to fill this gap. For this purpose, we derive a conceptual framework by synthesizing the literature on AI, including related fields such as ML.

The paper is structured as follows. First, we provide the methodology and outline the framework. Then we state the outcome of the literature review with details of the outlined framework. Then, we present the entire framework together with a discussion on key differences to other frameworks that originate from the different nature of AI (compared to IT). Next, we provide a discussion and future work. In particular, we show how theory development can be carried out using our framework. We conclude with a summary and avenues for future research.

## Methodology

We began our study with a systematic literature search. While this yielded several interesting works, we found them insufficient to cover AI governance from a business perspective, lacking important areas related to governance and AI. Thus, we enhanced the systematic search with a narrative search (King & He, 2005) to fill gaps not addressed by articles from the systematic literature review. The narrative search was guided by approaching AI governance from two complementing perspectives (i) looking at governance frameworks on IT (Tiwana et al., 2013) and data (Abraham et al. 2019) with emphasis on business-relevant aspects and investigating to what extent each aspect in this framework was covered by literature, (ii) looking at key characteristics of AI that are likely impacting their governance and investigating to what extent these characteristics have been addressed. An overview of the process is provided in Figure 1.

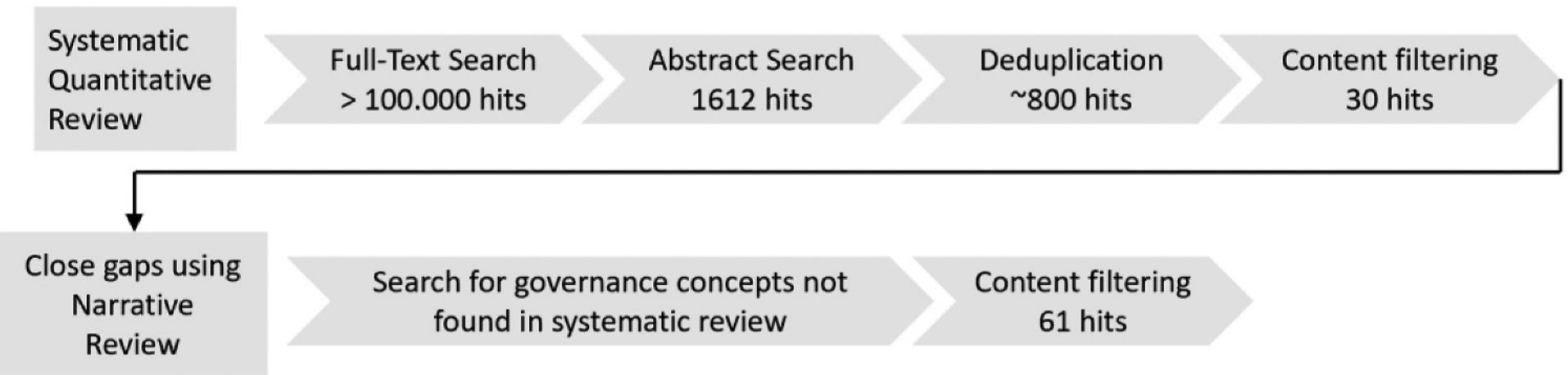

**Figure 1.** Overview of literature review process.

We started with a search covering keywords (Machine Learning OR AI OR Artificial Intelligence) AND (governance OR management) AND (compan* OR business* OR firm? OR organization?) on all available databases at our university, i.e., ProQuest, EBSCO, Google Scholar, Science Direct, Springer, and AISnet, restricting to peer-reviewed journals and English articles. Other prominent databases such as Web of Science and Scopus have been superseded by Google Scholar (Martín-Martín et al., 2018). The chosen databases also cover our key areas of interest, i.e., managerial works, works in computer science and works at the intersection, e.g., information systems. We considered articles published between 2011 and May 2021. A full-text search gave 30,000 hits on Springer and even more on Google Scholar. Thus, we further limited our search to abstracts. This gave a total of 1,627 hits (1,212 without governance and 415 without management). We estimate that about half of all the hits were duplicates. Articles were removed mostly based on their title and abstract. A large number of articles were non-AI specific (e.g., discussing also blockchain, Industry 4.0 or containing AI, which abbreviated another term) or not focusing on a single company (e.g., public policy articles) or unrelated to governance or management of AI (e.g., using AI to support management). For governance, this left us with 13 out of 415 articles. For management, many articles were domain-specific (e.g., manufacturing) and often discussed specific applications and challenges (e.g., predictive maintenance). While experiences of the management of specific cases are interesting, such articles are of limited relevance for a systematic literature review, leaving us with 17 articles, i.e., a total of 30 articles.

We found that these works on their own did not discuss multiple aspects related to governance as found in the context of IT, data and AI in a business context, e.g., how to value data. They also lacked tangible processes and procedures for ML. The identified works also did not allow to obtain a deeper understanding of key concepts in AI governance, allowing to set practical valid governance policies. Furthermore, a very large body of work is related to practical issues in ML that impact governance, e.g., ML governance and management, testing of ML models, ensuring fairness, explaining "black boxes." Thus, we also performed a narrative literature review (King & He, 2005). A narrative literature review puts the burden of an adequate search procedure on the reviewer. Our search was guided by synthesizing works identified in the systematic literature review and mapping them to existing governance frameworks (Abraham et al., 2019; Tiwana et al., 2013), taking into account key characteristics of AI such as its autonomy, ability to learn and obscurity. Parts of the framework that were not covered were explicitly searched for. Thus, we used a broad range of keywords, e.g., "AI design," "AI best practices," "AI safety," "AI strategy," "AI policy," "AI standards," "AI performance measurement," "data quality," "data valuation," "ML management," "ML process," "Model governance," "ML testing" accompanied by forward and backward search (Webster & Watson, 2002). Since the narrative literature review includes many search terms, we relied on a single engine. We used Google Scholar as it was shown to supersede prominent databases like Web of Science and Scopus (Martín-Martín et al., 2018). We removed the limitation to journals since new ideas are often presented first at academic conferences, and a significant body of works, particularly in computer science, only appear as conference articles. We focused on surveys if literature was abundant (for a keyword or during forward search). If there was a lack of peer-reviewed work, we also included gray literature, i.e., primary literature in the form of preprints from arxiv.org and reviews from MIT and Harvard. Other disciplines, namely computer science, frequently cite preprints on arxiv.org, which applies a moderation process for basic quality assurance, but no peer-review system. Pre-prints were only considered after our quality assessment. The main motivation for our process is (1) to include the most recent knowledge for the reader and (2) to cover the broad field of AI and governance from different angles. We believe this approach is necessary to ensure practical relevance and adequately discuss future research opportunities accounting for research that is likely completed soon. That is to say, pre-prints are typically scientific works currently or very soon under-peer review. Neglecting them puts a reader at risk of focusing on research gaps that are already in the process of being closed.

To analyze the identified works, we leverage existing frameworks on IT and data governance (as described in section Framework Outline). We used core, more generic dimensions shown in Figure 2. Individual works identified in the review process were first assigned into more generic categories shown in Figure 2. This was possible for all works. The generic categories contained sub-categories that stemmed from existing frameworks but were continuously expanded in the review process. We formed new categories for works that did not fit into the framework. Our final framework also dismisses aspects of prior frameworks (on IT and data governance) if they are not relevant in our context, as elaborated in the discussion of our final framework.

While most of the literature analysis was done by one researcher, the resulting framework was discussed by all coauthors.

## Framework outline

We first describe the two existing frameworks that built the foundation of our framework. We defer a detailed discussion of differences to the end of the paper, presenting the final framework (Figure 8).

Abraham et al. (2019) presented a framework for data governance. It covered only the data dimension, which we expand with concepts that arise or allow for different, more specific treatments in the context of AI such as data valuation, bias, and (concept) drift. For data governance, we focus on differences in governance resulting from a company using data for AI or other purposes. For the latter, we refer to Abraham et al. (2019) instead. The model and system dimensions are of equal importance to the data dimension. The dimensions and concepts are largely a synthesis of multiple existing works stemming from non-technical articles on AI governance and more technical articles on ML. We also state antecedents and consequences. Antecedents describe internal or external factors that impact AI governance. These include internal factors that are prevalent during the adoption of AI governance, as well as internal and external factors that are at best marginally influenced by governance initiatives such as legal requirements. Consequences describe outcomes of AI governance that can be directly or indirectly related to a corporation's goals. Figure 2 provides an overview of the research framework. Figure 8, toward the end of the article, includes additional details.

Abraham et al. (2019) built upon Tiwana et al. (2013), which addresses the who, what, and how of IT governance. In this work, for the who dimension, we primarily emphasize the firm and project dimension, only touching upon the governance of large units of analysis such as ecosystems. We recognize that AI being a composition of data, models, and systems, lends itself naturally to govern both IT artifacts (models, systems manifested as software code) and content (data) found in the "what dimension" of Tiwana et al. (2013). As shown in the following sections, we refine Abraham et al. (2019) by defining AI governance as the union of data, model, and systems governance (see, Figure 2).

## Literature review

This section elaborates on AI governance based on the reviewed papers. First, we define the scope in the form of an AI governance definition more clearly before employing the structure of the conceptual framework shown in Figure 2. Each dimension of the conceptual framework is refined with findings from literature. We start with the description of governance mechanisms and afterward move on to subject scope, organizational scope, and targets.

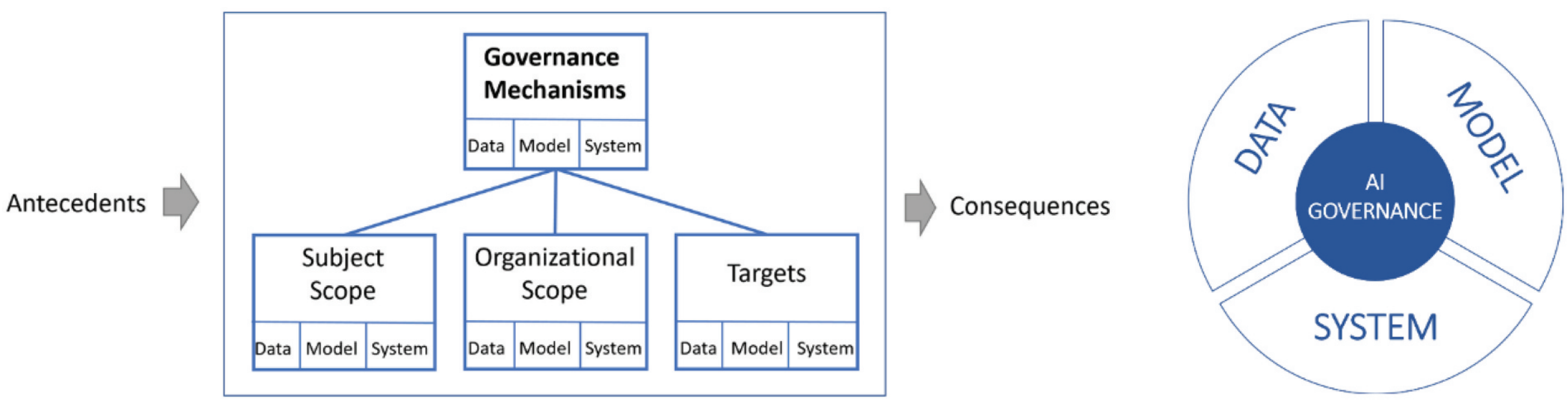

Figure 2. Our perspective on AI governance (left) and the core of our conceptual framework (right).

### *AI governance definition*

To clarify the scope of this work, we state a working definition of the key term "AI governance for Businesses." It combines the well-defined term "corporate governance" and "AI," which shows more variance in the literature. We propose the following definition:

**AI governance for business is the structure of rules, practices, and processes used to ensure that the organization's AI technology sustains and extends the organization's strategies and objectives.**

Aligned with Abraham et al. (2019), we emphasize the following six parts of AI governance:

- fostering collaboration across functions
- structuring and formalizing AI management through a framework
- focusing on AI as a strategic asset
- defining how and who makes decisions
- developing supporting artifacts (policy, standards, and procedures), and
- monitoring compliance.

AI aims at building "machines that can compute how to act effectively and safely in a wide variety of novel situations" (Russell & Norvig, 2020). ML can be seen as a branch of AI with the narrower goal of learning from data. Thus, ML denotes the models and processes for building and testing models by learning from data. AI aims at building an AI system consisting of an ML model and typically other components, e.g., graphical user interface and input handling for an AI (software) system or physical elements such as those found for a robot. The ML model might be seen as the brain of the AI, i.e., it is responsible for making decisions.

The term AI differs from the one of ML. It is not only broader regarding its goal but also in terms of the techniques used to achieve intelligent behavior. While ML focuses on learning from data, AI includes non-learning -based techniques, e.g., logical deduction (Russell & Norvig, 2020). We use a data-driven lens of AI based on the observation that most existing AI techniques, e.g., deep learning, incorporate some form of learning from data given as a set of samples.

Data representing experience in the form of samples might be based on an exploration of an environment by an agent using reinforcement learning (RL; Russell & Norvig, 2020) like in "Alpha Go Zero" (Silver et al., 2017). In this case, no domain knowledge in the form of data is available, but the agent generates it through exploration. We do not discuss RL in detail, mainly because businesses reap most benefits with supervised models and RL is less commonly used in practice (Witten & Frank, 2017).

From a governance perspective, it is appealing and natural to define AI in the following three areas: data, model, and (AI-)system since they relate to existing governance areas. Furthermore, they build on top of each other. AI governance refers to data and IT governance, covering both systems (commonly a large software code basis) and models (typically a small code basis). In contrast, model governance is a new area that comprises the governance of relatively small software codes (compared to traditional software) that contain models and procedures for training, evaluation, and testing.

### *Governance mechanisms*

Businesses can make use of various mechanisms to govern AI. They consist of formal structures connecting business, IT, data, model, ML, and system management functions, formal processes and procedures for decision-making and monitoring, as well as practices supporting the active participation and collaboration among stakeholders (De Haes & Van Grembergen, 2009; Peterson, 2004). Following the literature on IT governance (De Haes & Van Grembergen, 2009; Peterson, 2004), we employ the separation between (a) structural, (b) procedural, and (c) relational governance mechanisms. An overview is given in Figure 3.

#### *Structural mechanisms*

Structural governance mechanisms define reporting structures, governance bodies, and accountability (Borgman et al., 2016). They comprise (1) roles and responsibilities and (2) the allocation of decision-making authority. While for data governance, there is a rich set of literature (Abraham et al., 2019), literature is relatively sparse for AI governance. Ho et al. (2019) briefly discussed the responsibilities of a chief data officer (CDO) and chief information officer (CIO) in the context of image analysis in healthcare. Using a hub that centralizes responsibilities such as talent recruitment, performance management, and AI standards, processes, and policies has also been set forth (Fountaine et al., 2019). In the adoption phase, establishing a center of excellence has been suggested (Kruhse-Lehtonen & Hofmann, 2020).

Developing AI systems is a multi-disciplinary endeavor (Tarafdar et al., 2019). Thus, establishing an (interdisciplinary) AI governance council that has been advocated for AI in healthcare (Reddy et al., 2020) might be needed to handle the complex interrelation between model outputs, training data, regulatory and business requirements. Executive sponsors also play an essential role (Pumplun et al., 2019). To what extent the executive sponsor controls and decides on performance goals might depend on the company's stage of adoption. For example, Pumplun et al. (2019) advocated that at least in the early phase of adoption, a dedicated AI budget without any performance obligations might benefit adoption. More specific roles related to model aspects are still subject to investigation. Serban et al. (2020) view the assignment of an owner to each feature as best practice.

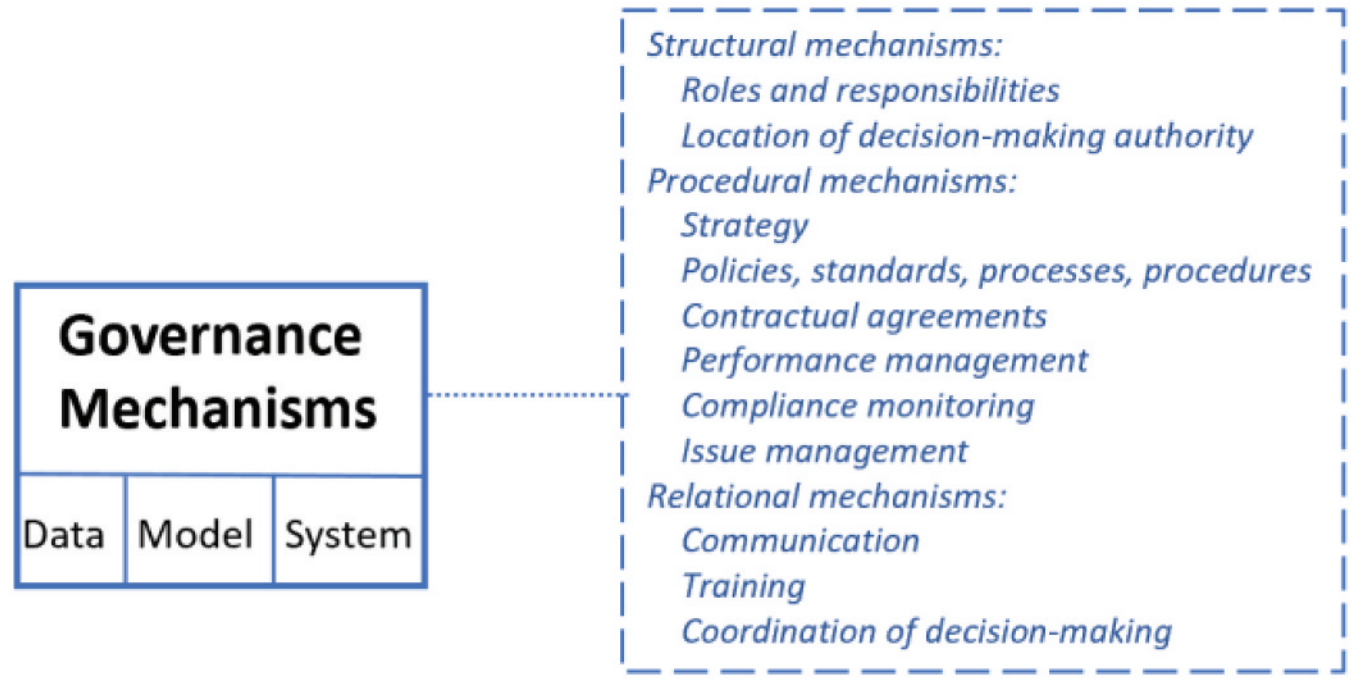

**Figure 3.** Overview of governance mechanisms.

### *Procedural mechanisms*

Procedural governance mechanisms aim to ensure that AI systems and ML models operate correctly and efficiently, they are held securely, and their operation meets legal and company internal requirements and policies concerning explainability, fairness, accountability, security, and safety. While ultimately, these goals must be fulfilled for the entire system and the ML model, explainability, fairness, and accountability are primary concerns of ML models. In contrast, safety is a more holistic trait, e.g., while an ML model might be deemed unsafe, safeguards such as non-AI-based backup components providing only reduced functionality might ensure the safety of the entire system. A "big red emergency button" might at least allow disabling the AI at any time, though the usefulness of this idea is subject to debate (Arnold & Scheutz, 2018). Procedural mechanisms also aim to ensure data, model, and system-relevant traits and targets, as shown in Figure 3. For data, model and systems individually and holistically, they comprise (1) a strategy; (2) policies; (3) standards; (4) processes and procedures; (5) contractual agreements; (6) performance measurement; (7) compliance monitoring; and (8) issue management.

The strategy denotes high-level guidance of actions based on strategic business objectives. The review by Keding (2020) investigated the interplay of AI and strategic management, focusing on two aspects (a) antecedents, i.e., data-driven workflows (data value chain and data quality), managerial willingness, and organizational determinants (AI strategy and implementation), (b) consequences of AI in strategic management on an individual and organizational level (e.g., human-AI collaboration, AI in business models). The work is focused on the use of AI for decision-making within a company, i.e., within a business intelligence context. There is work on several elements related to AI strategy, such as guiding principles by academics and practitioners (Iansiti & Lakhani, 2020; Kruhse-Lehtonen & Hofmann, 2020; Pichai, 2018; Smit & Zoet, 2020).

AI policies provide high-level guidelines and rules. Organizations use AI policies to communicate key objectives, accountability, roles, and responsibilities. There is active discussion in politics on AI policies. For example, in a white paper by the European Union (EU; European Commission, 2020), policy options have been set forth on how not to hinder the uptake of AI while limiting risks, and a subset of it is already part of a proposal (European Commission, 2021). On an organizational level, one source for policies might be best practices (Alsheiabni et al., 2020). Measurement in (public) AI policy is also an active field of debate and research, e.g., Mishra et al. (2020) discuss aspects of how to measure AI's societal and economic impact and risks and threats of AI systems.

While the development of standards concerning data has widely progressed (DAMA International, 2009), the development of AI standards is currently ongoing, e.g., Wellbeing metrics for ethical AI (IEEE P7010), benchmarking accuracy of facial recognition systems (IEEE P7013), fail-safe design for AI systems (IEEE P7009), ethically driven AI Nudging methodologies (IEEE P7008) (see, Cihon (2019) for a more comprehensive overview). In our framework, standards should ensure that data representations, ML models, and the architecture of AI systems, including their accompanying processes, are consistent and normalized throughout the organization. They should support interoperability within and across organizations and ensure their fitness for purpose (Mosley et al., 2010).

According to International Organization for Standardization's (ISO) Standard 9000, a process is a "set of interrelated or interacting activities, which transforms inputs into outputs" and procedures as a "specified way to carry out an activity or a process." In the context of governance, we view procedures and processes as standardized, documented, and repeatable methods. AI processes can be considered a fundamental element of a successful AI governance implementation. They relate to data governance processes (Abraham et al., 2019) and processes tied to ML and AI systems. ML model development relies commonly on elements of the cross-industry standard process for data mining (CRISP-DM; Wirth & Hipp, 2000). Coding guidelines in software engineering can also serve as a governance mechanism. Similarly, developing and enforcing model (and system) architectural principles, documentation, and coding guidelines with the goal of better composability and maintainability might also be a governance instrument for AI to foster interoperability (Stoica et al., 2017). Serban et al. (2020) view the documentation of each feature, including its rationale, as best practice.

Other proposals have been made with influences from software engineering (Amershi et al., 2019; Liu et al., 2020; Rahman et al., 2019; Watanabe et al., 2019). While Amershi et al. (2019) rely on surveys within a single company, Rahman et al. (2019) generalize from a case study describing the use of ML to correct transaction errors. However, there is still active discussion on which process is best (Liu et al., 2020). Palczewska et al. (2013) discussed model governance listing processes for model evaluation (e.g., review policies), control (changes and security, documentation), and validation. Testing procedures of ML systems received significant attention (Zhang et al., 2020), as well as assurance of ML (Ashmore et al., 2021; Hamada et al., 2020), i.e., executing an ML development process that delivers evidence for the model's safety. Furthermore, some of the works covered under processes might also exhibit a procedural character. Overall, there is a lack of research on model and AI procedures.

Contractual agreements between participating internal departments or external organizations might be related to data but also models and AI systems. Models contain information on the training data that might be extracted and abused in various ways, e.g., by competitors to reduce costs for data labeling. For example, consider a system that allows uploading an image and obtaining a classification. A competitor might use the system to label her unlabeled data. The competitor might use the resulting dataset to train a novel classifier and offer an "identical" service. Contracts also play an important role in reducing liability risk by clearly specifying operational boundaries of AI. AI might malfunction under circumstances that are non-intuitive and unexpected for a human. Therefore, operational constraints and limitations require explicit legal notices. Classical risk management approaches have also been suggested for AI (Clarke, 2019). AI agency risk and the mitigation using business process management have been discussed on a conceptual level (Sidorova & Rafiee, 2019).

Compliance monitoring tracks and enforces conformance with regulatory requirements and organizational policies, standards, procedures, and contractual agreements. Prominent regulations include the European General Data Protection Regulation (GDPR), which touches upon data and models, e.g., it grants the right to explanations of model decisions. GDPR is supported by extensive guidelines on compliance (Privacy Team, 2020). While there is some conceptual understanding of AI from a legal perspective (Scherer, 2015), and there are multiple non-legal statements from governments, e.g., European Commission (2020), the body of regulations and, thus, the need for compliance monitoring is likely to grow. Audits allow for interrogating complex processes to clarify whether these processes adhere to company policy, industry standards, or regulations. Brundage et al. (2020) discuss AI auditing. They emphasize its need due to the complexity of AI.

Issue management refers to identifying, managing, and resolving AI-related issues. For data issues (Abraham et al., 2019), this includes processes for the standardization of data issues and their resolution, the nominating persons for issue resolution, as well as an escalation process. For model and AI systems, little information is available.

### *Relational mechanisms*

Relational governance mechanisms facilitate collaboration between stakeholders. They encompass (1) communication, (2) training, and (3) the coordination of decision-making. Serban et al. (2020) express the need to communicate within an interdisciplinary ML team and suggest utilizing a collaborative development platform. Training procedures for data scientists have been put forward in Biomedicine by Van Horn et al. (2018).

They analyzed a rich set of training resources and conducted training events on applied projects. Mikalef and Gupta (2021) stated that technical and business skills are essential AI capabilities. Thus, training employees is critical. While training might often refer to developing skills to leverage AI, it might also refer to training employees whose tasks and responsibilities might be automated or augmented by AI to mitigate negative impacts (Kaplan & Haenlein, 2019). Communication might help reduce employees' fear, e.g., by signaling a company's intention to use AI as augmentation of workers rather than a replacement (Kaplan & Haenlein, 2019). Overall, little is known for ML models and AI systems governance in contrast to data governance (Abraham et al., 2019; Borgman et al., 2016).

## Subject scope

This dimension pertains to what is governed. It covers IT artifacts (models and systems) and content (data), as shown in Figure 4.

### Data subject

Data is the representation of facts using text, numbers, images, sound, or video (DAMA International, 2009). While Abraham et al. (2019) provide an overview of the data scope, we emphasize that the type of data impacts the model selection, e.g., structured or tabular data vs. unstructured data such as text and images. For example, deep learning models work well with large unstructured data but have been relatively less successful for tabular data. An essential characteristic of data is its sensitivity level: Is it personal or non-personal data? Personal data, i.e., data that relates to humans, underlies different regulations, e.g., GDPR enacted in the European Union in 2018. A secondary reason for the separation is that human data often come with different characteristics with respect to quality and costs. These are strongly reflected in ML and system properties. For instance, GDPR grants the right to explanation to individuals for automated decisions based on their data. Such explainability processes are less of a concern for models based on non-personal data. Reddy et al. (2020) set forth a governance model for healthcare based on fairness, transparency, trustworthiness, and accountability. Janssen et al. (2020) emphasized the importance of a trusted data-sharing framework.

### Model subject

Supervised learning uses labeled training data. That is, each sample $X$ is associated with a label $y$. Unsupervised learning consists of unlabeled training data. It typically addresses tasks like organizing data into groups (clustering) or generating new samples. While adaptivity might be a requirement in a dynamic environment, online systems, which continuously learn from observed data, are more complex from a system engineering perspective with multiple unresolved challenges (Stoica et al., 2017). System evolution must be foreseen and managed, which is not needed for offline systems that do not change. In the past, such dynamic systems have gotten "out of control" after deployment, e.g., Microsoft's Tay chatbot had to be shut down after only a few hours of operation due to inappropriate behavior (Mason, 2016). Finally, an important concept is that of transparency of ML models. Transparent models are intrinsically human understandable, whereas complex black-box models such as deep learning require external methods that provide explanations that might or might not suffice to understand the model (Adadi & Berrada, 2018; Meske et al., 2022; Schneider & Handali, 2019).

### System subject

Autonomy refers to the decision-power of AI. It is commonly discussed in the context of agents (Franklin & Graesser, 1996). Autonomy impacts legal responsibilities and liability. The discussion of autonomy relates closely to human-AI collaboration. For example, a fully autonomous AI might decide on the treatment of a patient without any human review or possibility of intervention. At the same time, an assisting AI might only suggest a treatment, and the decision is left to a human. AI might augment jobs or tasks, requiring close interaction with human collaborators or fully automate them (Davenport, Thomas & Kirby, 2015). There are widespread concerns about the future of work in the face of AI (Jarrahi, 2018) and policy recommendations by regulatory bodies on the designation of decision control (European Commission, 2020). Human-AI collaboration also plays an important role from a strategic perspective (Keding, 2020). A survey of human-AI interaction (Rzepka & Berger, 2018) categorized works into the topic of users, system, interaction, task, context, and outcomes. Relevant system characteristics also cover the appearance of AI and gestural and conversational behavior. These characteristics also play a prominent role in assisting digital AI agents (Maedche et al., 2019) and social robots (Nocentini et al., 2019). Guidelines for human-AI interaction have also been set forth by Amershi et al. (2019). The guidelines serve for initial usage, interaction, behavior in case of errors, and behavior over time. They can also be categorized as informing users, managing decision control, and collecting and leveraging user feedback. AI safety refers to the property of not causing danger, risk, or injury (Amodei et al., 2016). Highly safety-critical systems typically put human lives at risk, such as autonomous cars. The environment in which an AI system operates strongly impacts its properties' relevance and required levels, such as security and robustness. Dynamic environments are subject to rapid and large changes, while static environments do not evolve. Dynamic environments often require online learning, i.e., highly adaptive models that continuously learn.

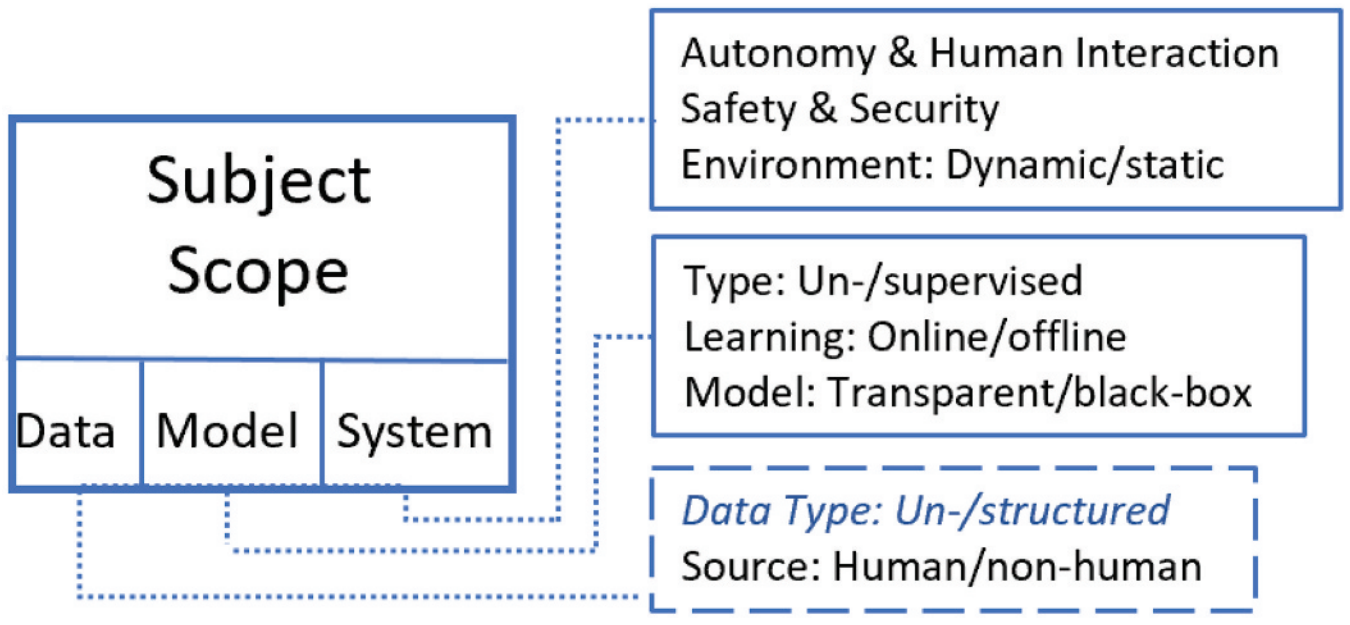

**Figure 4.** Overview of subject scope.

## Organizational scope

The organizational scope defines the breadth of governance in terms of entities involved. Following Tiwana et al. (2013), it consists of the intra- and inter-organizational scope as shown in Figure 5. The intra-organizational scope refers to governance within a single organization whereas the inter-organizational scope refers to governance between multiple organizations. For AI governance in an intra-organizational scope, this relates to quality and integrity of data, models, and systems on a project level. At the same time, the firm-level considers the demands of multiple stakeholders dispersed within the company. While these aspects are well-studied for data governance (Abraham et al., 2019), more research is needed for model and system governance. Co-development among multiple partners, including public entities such as universities or private entities such as individuals in hackathons, are seen as ways for successful AI development (Kaplan & Haenlein, 2019). Such collaborations require appropriate governance. Aside from direct co-development, existing knowledge or artifacts such as models are often freely available, e.g., open-source deep learning modes or open-access research papers. Models might be trained using publicly available data and code or even be pre-trained on a large dataset and only adapted for a company's specific need, e.g., using transfer learning on a small, company-owned dataset. While open source initiatives and platforms are not new, AI arguably further pushed the extent of these initiatives, e.g., most designs of deep learning models for image tasks or natural language processing (NLP) originate from publicly available models. They are often obtained from repositories such as GitHub or already integrated into frameworks such as Tensorflow or Pytorch. Governance of open-source code or platforms is a topic on its own (Halckenhaeusser et al., 2020). Their governance poses new challenges for the governance of AI in organizations and raises questions how to ensure that such models are compliant with corporate policies and how to mitigate liability issues. This is particularly relevant because models might suffer from different forms of biases and might malfunction in non-predictable ways. Also, their inner workings are hard to assess due to their black-box nature.

## Targets

Governance programs address goals in one or multiple areas, as summarized in Figure 6. While we discuss data, models, and systems separately, most works discuss only one aspect. Siebert et al. (2020) propose guidelines for assessing qualities of ML systems, discussing criteria related to data, model, environment, infrastructure, and system. We subsume their model criteria focusing on metrics relevant to governance.

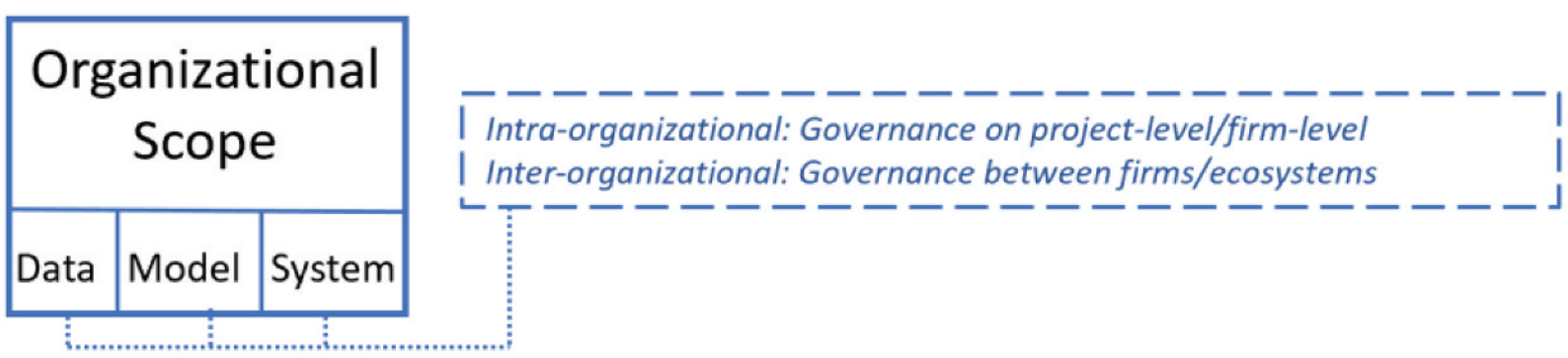

Figure 5. Overview of organizational scope.

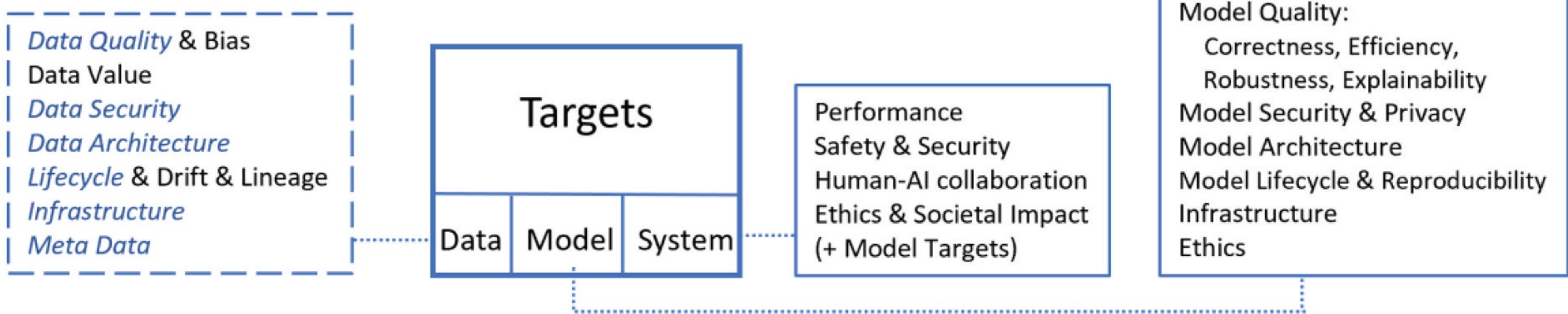

**Figure 6.** Overview of targets.

### *Data targets*

Data governance aims to maximize data value at minimal cost (Abraham et al., 2019). However, data valuation is currently not part of data governance and AI (Abraham et al., 2019). The value of data is difficult to estimate since potential uses of data and the resulting benefits are hard to foresee. Pricing data on data marketplaces or quantifying the right amount of data to purchase is non-trivial (Jiao et al., 2018; Pei, 2020). For example, if data comes for free and is not traded, e.g., in the form of trace data associated with business transactions, its value is often not of primary concern due to low storage costs and data usage only inside the business. However, data valuation gains in relevance if the acquisition of data comes with costs, e.g., if data has to be labeled by humans as part of the construction of a dataset for an AI system or if data requires costly processing, such as manual cleansing to raise data quality. Some general principles qualitatively describe the value of data for ML, such as that the value of data increases with quality and exhibits diminishing marginal utility. Some works allow quantifying the impact of a data sample on the correctness of an ML model, e.g., using XAI techniques (Ghorbani & Zou, 2019; Meske et al., 2022) though they are often computationally demanding. These techniques answer the question: "How much accuracy is lost (or gained) if we do not utilize a specific data sample for training a model?." This provides a solid basis for data valuation of single samples but also datasets. More proactively, active learning (Aggarwal, 2015) allows for determining data samples that are giving most improvements for a given model. In ML, training data serves the only purpose to increase model performance. The value of a performance increase of an ML model might often be better and easier to quantify than that of data in general (without knowing any context on its usage). For example, improvements can be valued by increasing revenue from recommendations for a recommender system in an online shopping setting. This allows to more easily assess the value of data for a particular system – down to a single sample. Quantifying data value at least on an aggregate level is highly desirable to make informed investment decisions. The value of data also changes over time. While there might be multiple reasons for this, concept drift is a major concern in the field of AI that reduces data value.

Concept drift refers to how the data distribution changes over time (Gama et al., 2014; Lu et al., 2018). Concept drift implies that the data used to train the ML model does not capture the relationship that the model should capture. Concept drift typically gets more severe over time. One can distinguish between real and virtual drift (Lu et al., 2018). Real drift refers to a change in the classification. It captures the scenario that a particular action might be deemed legal up to a regulatory change and, from then on, illegal. Virtual drift denotes a change in distribution of samples between training and operation. For example, a change of distribution might occur if a system used and designed for adults is used increasingly by younger people in somewhat different ways. Janssen et al. (2020) proposed monitoring the outcomes of algorithms. However, ideally, changes in data are detected before algorithms are impacted, as discussed in literature on concept drift, e.g., Zhang et al. (2020).

Data quality denotes the ability of data to meet its usage requirements in a given context (Khatri & Brown, 2010). The relevance and handling of data quality might differ for AI systems compared to business intelligence applications or other forms of analytics. In the context of AI, better data quality is always preferable from a model performance perspective. However, improving data quality also comes with high costs. Given that state-of-the-art ML models have shown some robustness to certain forms of data quality issues (Frénay & Verleysen, 2013; Nigam et al., 2020; Song et al., 2020), the preferred data quality from an economic perspective might be a revenue tradeoff between increased model performance and costs for data cleansing. There also exist techniques to filter "bad" data that might even deteriorate model performance when used for training – see, Brodley and Friedl (1999) for pioneering work and Ghorbani and Zou (2019) for a recent technique. Therefore, data filtering techniques might replace data cleansing given that a large amount of mostly accurate data exists.

Another data quality aspect related to fairness of ML systems is bias. Data is biased if it is not representative of the population or phenomenon of study. Biased data might lead to biased ("unfair") ML models (Verma & Rubin, 2018) that violate regulations.

Data lineage includes tracking its origin, where it moves and how it is processed. In particular, in AI, this includes connecting data, pre-processing, and models using the pre-processed data (for training; Vartak et al., 2016). The same data might be used in many different ML models, e.g., through transfer learning. Thus, proper governance should ensure that any updates, e.g., improvements in data quality, should be propagated adequately to all relevant stakeholders.

The data lifecycle represents the approach of defining, collecting, creating, using, maintaining, archiving, and deleting data (e.g., Khatri & Brown, 2010).

This step further encompasses deriving data retention requirements from business, regulatory, and accountability demands.

### Model targets

Similarities and differences of data and model targets can be seen by comparing the targets for data and model in Figure 6. Model quality relates to a number of different concepts, which might be extended and less relevant depending on the application. Security and privacy relate to two separate matters: Security is a target of a model, e.g., "Is the model secure against attacks during operation?" and an attribute related to handling the model, e.g., "Is the model stored securely with proper access rights (in the development environment)?" For data, only the second concern applies. While metadata is highly important to establish the semantics of data, for models, this is typically encoded in the artifact itself, i.e., the model definition. However, metadata on models such as who authored or changed the model is relevant to ensure accountability.

ML models are commonly assessed and tested for the following dimensions (Zhang et al., 2020): correctness, efficiency, fairness, interpretability, robustness, security, and privacy. Zhang et al. (2020) also list model relevance, which measures to what extent a model overfits or underfits the data as testing property. However, it is implicit that relevance of a model increases with correctness of the model.

Correctness of an ML model measures the probability that the ML system works as intended. For classification tasks, it is often reported as the percentage of correctly categorized samples of a classifier (accuracy) for a test dataset. It is often seen as the key determinant for model performance judgment. Robustness defines to what extent an ML model can function correctly in the presence of invalid inputs (Zhang et al., 2020). For instance, it allows assessing to what extent concept drift is a concern. It is also related to security since common attacks on ML models exploit their non-robustness. More generally, security is the resilience against negative impacts caused by manipulations or illegal access to ML models. Privacy, which is also a security concern, relates to information confidentiality. It measures to what extent an ML system can maintain private data, e.g., if private data be extracted from a model trained on private data.

Decisions of an ML model might not be in accordance with accepted or legally binding "fairness" definitions. There is no consensus on what fairness comprises, and multiple, potentially even conflicting definitions exist (Verma & Rubin, 2018). However, there is widespread acceptance among the public and legal authorities that decision criteria denoted as protected characteristics (Corbett-Davies & Goel, 2018) such as gender, religion, familial status, age, and race must not be used. While fairness might often be traced back to data, it should primarily be assessed in ML models for two reasons: (1) decisions are made by models (not by data); (2) the impact of data bias on the fairness of decisions of an ML model is difficult to anticipate. Data might be biased, and a model's decisions are not (that much) or the other way around. For illustration, there are methods to counteract fairness issues on a model level (Du et al., 2020), i.e., a fair model might be obtained using such mechanisms even in the face of data bias.

Explainability relates to understanding model behavior. A common definition states that it provides the degree to which a model's decision or the model itself can be understood. Understanding might improve trust of end-users as well as help in model testing and validation by engineers. Efforts of explainable artificial intelligence (XAI) are also driven by legislation, e.g., GDPR granting rights to individuals for an explanation involving personal data. XAI is still subject to research within the information systems community (Adadi & Berrada, 2018; Meske et al., 2022; Schneider & Handali, 2019) and other disciplines (Das & Rad, 2020). Efficiency of an ML system relates to the computational effort to construct a model or perform a prediction. In a practical setting, it is often approximated by the time it takes a computer to complete either of the two aforementioned tasks, i.e., model construction and prediction. It is also a key focus of computer science, often estimating worst-case behavior measured in abstract operations. In particular, if AI systems are to be deployed on resource-constrained systems such as mobile devices, efficiency of prediction is a primary concern.

In contrast, efficiency of model construction might limit productivity of data scientists and ML engineers and impact overall project duration. ML models might take weeks to train on ordinary personal computers, particularly if multiple models should be assessed. In contrast, distributed training using specialized hardware (Verbraeken et al., 2020) of the very same model might take only a few minutes. Thus, infrastructure relates to the software and hardware available to train, test and deploy ML models. Enabling parallel training of ML is considered a best practice (Serban et al., 2020). It relates to efficiency and scalability of AI systems to a growing number of predictions as well as to growing complexity and a growing number of models. While there are differences in hardware used, e.g., GPUs and scheduling jobs for ML (Mayer & Jacobsen, 2020), overall governance aspects are similar to non-AI specialized IT.

According to Sridhar et al. (2018), solutions for production ML should track provenance/lineage, ensure reproducibility, enable audits and compliance checks of models, foster reusability, handle scale and heterogeneity, allow for flexible metadata usage. Sridhar et al. (2018) go beyond these recommendations and provide an architecture and design to facilitate "model governance." The complex nature of AI has even led to the idea of employing third-party audits on internal development (as a common routine; Brundage et al., 2020).

Model management is an emerging topic (Vartak et al., 2016), with an increasing number of tools becoming available, e.g., Chen et al. (2020) provide a recent example. While it is often developer-focused, it also addresses points relevant to model governance, managing modeling experiments, enabling reproducibility, and supporting sharing of models. Model management comes with challenges on a conceptual, data management and engineering level (Schelter et al., 2018). These challenges include reproducibility issues, deciding when to retrain a model, lack of abstraction for the whole ML pipeline (such as the Structured Query Language [SQL] for data), access to model metadata, multi-language codebase. Challenges of ML management have also been expressed relative to software management leading to three main challenges (Amershi et al., 2019): (1) discovering, managing, and versioning is more complex for ML (than for data or software only), (2) model customization, and reuse require skills not found in software teams and (3) AI components are more difficult to be handled as separate models due to their entanglement and error behavior. A model management system is supposed to track the data pipeline, i.e., to record which data and transformations have been used to build and test a model. This is analogous to versioning using code repositories for classical software. It should also support the sharing of models and, potentially, levels of abstractions of ML frameworks.

Model management systems might furthermore support testing, monitoring, and analysis of models. Serban et al. (2020) promote software engineering best practices for ML grouped into six classes, i.e., data, training, coding, deployment, team, and governance. Several practices relate to model management, recommended testing practices. Palczewska et al. (2013) discussed model governance based on data governance tailored toward predictive models for toxicology in food and related industries. While they miss some aspects of contemporary model targets such as fairness and explainability, they discuss an information system for model management.

### *System targets*

The overall system is commonly assessed using performance metrics that strongly depend on model performance and encompasses other metrics such as system usability. Systems that collaborate with humans are of particular interest. They pursue different goals, e.g., the collaboration of AI and humans can either substitute or augment existing work or assemble new dynamic configurations of humans and AI that provide a single, integrated unit (Rai et al., 2019). For systems interacting with humans, data guardianship, system accountability, and ethics in decision-making as well as security and privacy need to be considered (Rai et al., 2019). For example, a company might pursue the objective to minimize risks due to accountability resulting from AI decisions by augmenting user tasks rather than substituting, as is observed in a slow transitioning from assisted driving to autonomous driving (Bellet et al., 2019). AI safety (Hernández-Orallo et al., 2019) exhibits a rich set of characteristics, e.g., describing an operator's ability to intervene during operation, disposition to self-modify (autonomy), explainability of goals and behavior. It also relates to model robustness.

Furthermore, the system ultimately makes decisions based on model outputs. Thus, security and safety mechanisms such as a "big red emergency button" to disable the AI can be incorporated on a system level. AI ethics and implications have been extensively discussed in academic literature (e.g., Floridi et al. (2018)), and increasingly by institutional bodies (e.g., European Union (2020)). AI governance can aim to realize AI opportunities from a societal perspective, such as cultivating social cohesion, increasing societal capabilities, and limiting risks such as devaluing human skills and reducing human control (Floridi et al., 2018).

## Antecedents

Antecedents are the external and internal factors that influence the adoption of governance practices (Tallon et al., 2013, p. 143). We mainly synthesize factors of data governance from Abraham et al. (2019). We also investigated literature that discusses factors related to AI adoption (Chui & Malhotra, 2018; Fountaine et al., 2019), readiness (Jöhnk et al., 2021) and capability (Mikalef & Gupta, 2021) as well as antecedents specific to the use of AI-as-a-service (Zhang et al., 2020). We added existing AI development procedures and AI capability as (internal) antecedents based on these works. The rationale is that if a company has good development procedures and high AI capability, it must handle many concerns relevant to AI governance, such as data and model qualities. As such, different governance mechanisms might depend on AI capability. In particular, AI-business alignment might be relevant. It relates to its readiness based on harmonized and AI-aligned processes related to decision-making (Jöhnk et al., 2021). Furthermore, governance might also consider existing AI (development) procedures. Other internal factors relate to strategy, resources, and culture. Common factors such as top management support and financial budget also play a vital role (Abraham et al., 2019; Jöhnk et al., 2021). IT in general and enterprise architecture specifically have also been shown to be relevant as antecedents for data governance and AI adoption (Stecher et al., 2020). Since models build on data, collecting or accessing large amounts of high-quality data is essential. Cultural aspects relate to innovativeness, willingness to change, risk appetite, and collaborative work (Gupta and Mikalef, 2021; Jöhnk et al., 2021). Collaboration emphasizes the integration of different perspectives such as business domain (data), model and systems, and skillsets (technical and business) across organizational units.

External factors constitute legal and regulatory requirements relating to data such as GDPR (Privacy Team, 2020) or models and systems, e.g., effectively enforcing human agency for self-driving cars due to liability concerns. While AI is seen as a disruptive general-purpose technology (Magistretti et al., 2019), current AI is also deemed "narrow AI" due to its success for specific tasks only, i.e., its inability to show general intelligence. Thus, not surprisingly, AI deployment is industry and application-dependent (Chui & Malhotra, 2018). Customers must be ready for AI, and they must understand, e.g., its limitations, such as lack of transparency (Jöhnk et al., 2021). Customers are one of the stakeholders to be governed. As such, their readiness impacts governance.

## Consequences

We categorized the consequences of AI governance by extending Abraham et al. (2019) and Tiwana et al. (2013) into performance effects, risk management, orchestration, and, additionally, effects on people. It is worthwhile to emphasize effects on people due to the widespread discussion of AI and human interaction within the organization, i.e., employees and AI, and of the organization with outsiders, i.e., customers and AI systems of the organization.

It is also possible to dissect the impact of AI according to the subject to be governed, e.g., data, model, system, and stakeholder. Abraham et al. (2019) presented benefits of data governance on performance and risk management. Brundage et al. (2020) presented data governance intending to achieve trust. AI governance aims to increase AI capabilities derived from resources such as data, technology, human skills, and organizational aspects (Gupta et al., 2021). AI capabilities have contributed positively to firm performance and creativity (Gupta et al., 2021; Amabile, 2020). Furthermore, the use of AI (as a service) has also been shown to have a positive impact on firm performance (Zapadka et al., 2020).

An outcome of AI governance on employees, e.g., by prescribing adequate training for employees, might be the mitigation of negative automation effects. Impacted performance outcomes could be profitability, growth, stability, efficiency, or survival of a company. Organizational properties that are probably impacted by governance include efficiency, effectiveness, and flexibility. Even more than IT governance, AI governance is likely impacting orchestration, e.g., innovation partitioning. AI competencies and business knowledge are commonly separated, e.g., among departments, platform ecosystems, or cloud providers (offering ML services). Both entities being specialized differently might contribute to innovation.

## Final framework

Our work treated AI governance based on existing IT and data governance knowledge, seemingly viewing it as an evolution thereof. In particular, we used the IT governance cube of Tiwana et al. (2013), grounding our framework on the three general dimensions relating to what, who, and how is governed. While this seems to confirm the validity of the existing framework, we argue for an updated framework depicted in Figure 7 and a more refined form in Figure 8.

Tiwana's distinction of IT artifacts, content, and stakeholders seems only partially adequate. The notion of a model does not match the classical notion of an IT artifact, a piece of software or hardware that often contains data or information. A model is more declarative, and data determine its behavior. The exact content of a trained model is difficult to retrieve due to the black-box nature of AI, and it can only be said to be some function of the data. For many organizations, software and hardware are purchased externally and are fairly static. That is, updates are rare and costly since software must be explicitly programmed. In contrast, AI might be self-learning, and better-performing models might be obtained and serve primarily internal purposes. From our perspective IT artifacts as defined by Tiwana in terms of hardware and software might be seen as infrastructure that serves to store and process data as well as to develop models and, potentially, deploy AI systems. We view data, models, and systems as very tightly coupled, though they do not have to be governed as one unit, i.e., they might be governed using different targets.

Furthermore, Tiwana et al. considered the IT architecture as a potential means to govern. While enterprise architecture is relevant for AI adoption (Stecher et al., 2020), architectures, once implemented, are difficult to change, making it difficult to use a governance instrument in a dynamic environment. In the context of models, a model's architecture is relevant. Still, the objective they optimize and the training data are also highly relevant, making the term architecture unsuitable as an overarching term. Therefore, we exclude architecture. In the context of AI, aside from structural mechanisms (relating to decision rights) and procedural mechanisms (relating to control and standards and practices), relational mechanisms such as coordination of decision-making and training are highly important, which are not prominently present in Tiwana's cube. These three types of mechanisms originate from Peterson (2004) and De Haes and Van Grembergen (2009).Figure 8. Concepts within the conceptual framework for data governance. Concepts from (Abraham et al., 2019) are in dashed boxes with concepts in blue and italics.

## Discussion and future work

This work views AI as systems learning from provided data and decomposed AI artifacts to be governed into data, models, and systems. Models differ from software implementing explicit human stated rules, since models derive their behavior in an opaque manner from data and require special XAI techniques to (partially) understand. They also pose potential ethical and security risks, e.g., due to biased decision-making and adversarial attacks. At the same time models are very dynamic and can easily be updated or might even evolve autonomously by (re)training on newly collected data. These factors combined with growing legislation pose challenges for businesses to achieve the goal of leveraging the full potential of AI while mitigating risks. We believe that thorough governance is a key aspect of accomplishing this goal. In our governance cube that can guide companies and researchers alike, AI artifacts (data, model and systems) are part of the "What is governed?" dimension that together with two additional dimensions, i.e., "How is governed?" and "Who is governed?" as well as antecedents and consequences define our framework for AI governance. Tiwana et al. (2013) proposed theory development through permutation (and rotation) in their IT governance cube. Scholars were advised to pick a cell of the cube and study a selection of antecedents and consequences. We also call for doing so for our AI governance cube since we have presented various new mechanisms for governance and highlighted differences in antecedents and consequences compared to IT governance.

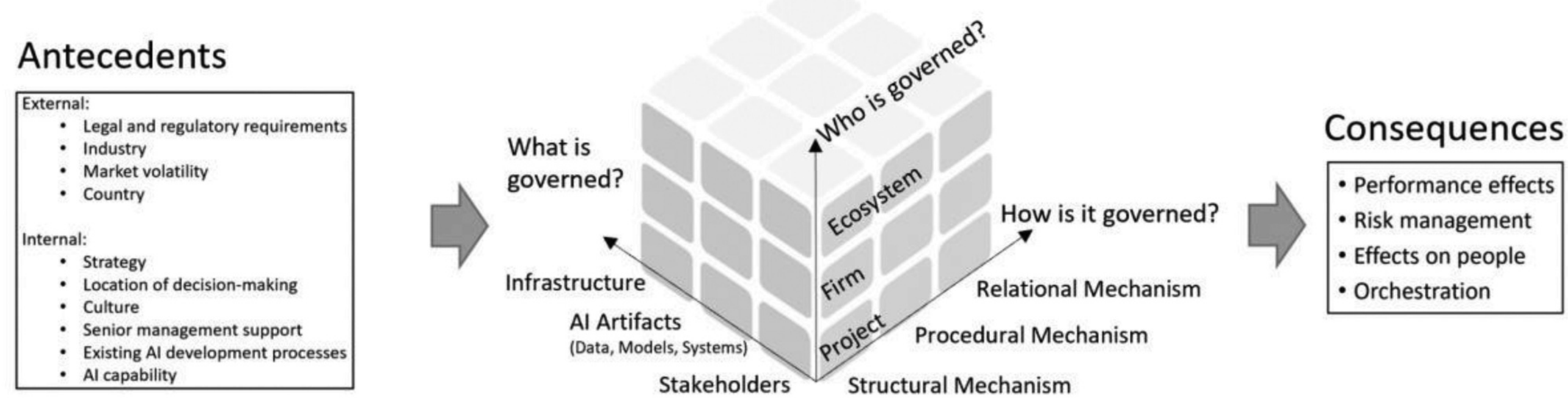

Figure 7. AI governance cube based on the IT governance cube (Tiwana et al., 2013). it can serve as a framework for theory building through rotation and permutation.

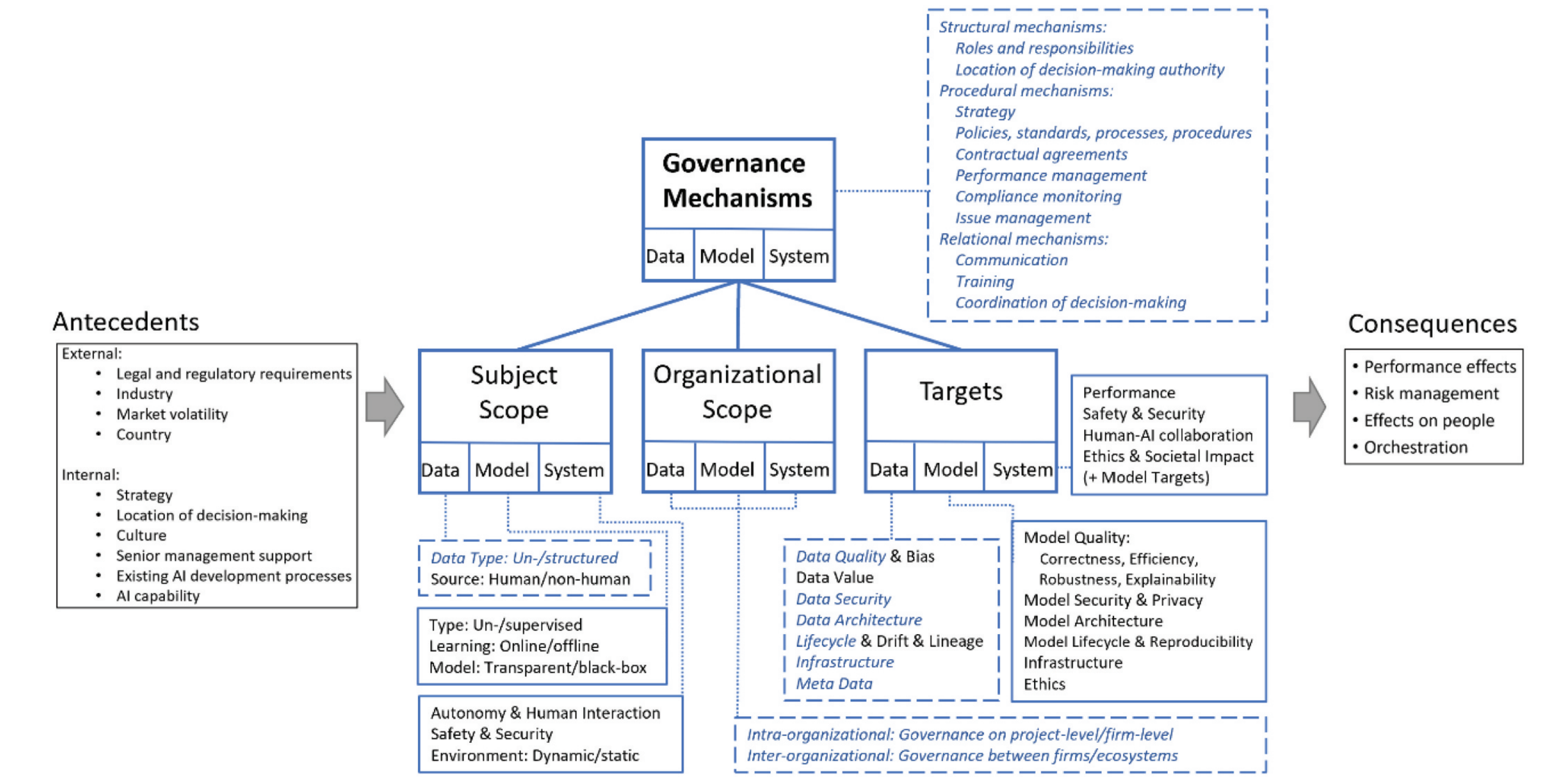

**Figure 8.** Concepts within the conceptual framework for data governance. concepts.

Many cells of the AI governance cube require more research and as AI is quickly evolving adjustments to governance are also likely needed. The future of AI might also bring a shift from current "narrow" AI toward artificial general intelligence (AGI). AGI will probably be even more autonomous optimizing prespecified goals without being given data. That is, it might derive its behavior by learning from her own collected data (rather than relying on given data) like agents in reinforcement learning as seen in groundbreaking systems like AlphaGo. This might bring about new challenges, such as the correct specification of goals and, in particular, adequate constraining of "allowed" solutions to achieve these goals. For example, ethical behavior must be ensured, i.e., by forbidding AI to deceive or endanger people to accomplish a goal. We hope and believe that despite the unpredictable evolution of AI, our framework will endure the test of time with some possible extensions. For example, we believe that data, models and systems will remain key AI artifacts, possibly enhanced by other artifacts such as the aforementioned specification of goals and constraints for AI. That is, even if data is collected and generated automatically, it is still of value and relevant for system understanding, e.g., the costs for training AlphaGo by learning from 29 Mio. played games are estimated to be more than 30 Mio. US dollars.[1]

Future developments of AI might shift the focus of governance – in particular, related to security, safety, and human-AI collaboration. Also, concerns with respect to ethics that are only briefly discussed in this work might become even more practically relevant. Already today, many aspects of AI governance remain non-specified and require more attention. While we highlighted many of these throughout the manuscript, we aim to contribute using several practical proposals, whose validity has yet to be verified:

The first set of proposals relates to structural mechanisms, i.e., roles. We propose to mimic the general structure of data governance (Abraham et al., 2019), however, with some alterations based on the following rationale: Both data and models share some characteristics that bear relevance for governance. Both should be shareable and employable in various areas of a business (Serban et al., 2020). Both undergo a lifecycle (often coupled). Both might be subject to audits and require accountability and testing. They also differ in various ways: Most production quality models undergo an engineering, training, and tuning process, which requires extensive technical knowledge. Data undergoes no processing aside from comparably simple transformations (cleansing, pre-processing) and computing KPIs, e.g., for data quality assessment. Data might also be used for AI unrelated systems. We suggest also specifying separate roles for ML development and ML (and AI) testing. This is also done for software to reduce the risk of bugs. Due to the prominent nature of AI risks, one might also introduce an AI risk manager that might, for example, veto against the rollout of an AI system.

With respect to infrastructure, we propose separating IT infrastructure for developing, testing, and deployment AI systems and ML systems. The reasoning is similar for software systems. Minimizing the number of people by carefully choosing access rights who can get a hold of sensitive data on a production server improves security. It reduces, e.g., the risk of privacy violations of customers. Separating systems also ensures that issues during development that might lead to outages of development infrastructure (e.g., running out of memory, utilizing all CPUs and GPUs) do not impact availability of deployed services. Furthermore, development hardware might be chosen to be of lower quality (and therefore be cheaper) since (short) outages are typically less of a concern as for a production environment.

Furthermore, future work might also focus on stakeholders such as employees and customers, which pertain to a special role, which could be studied in depth on its own. For once, they interact with an AI system. Therefore, they are exposed to risks associated with AI, e.g., customers might be exposed to unexpected behavior for inputs resulting in danger or, if AI is used for automation, it might cause job loss or require increased emphasis on continual learning to cope with change (Kaplan & Haenlein, 2019). Employees and customers might also provide data used for training models, often creating tension between their rights (e.g., related to data protection) and a business's intention to collect their valuable data. They might also have reservations due to trust issues (Kaplan & Haenlein, 2019). Furthermore, the development of models and AI systems often involves an interdisciplinary team requiring both technical and business skills (Gupta & Mikalef 2021). At least for some time to come, governments and society alike show a high level of interest in AI, making them stakeholders. They contribute to the discussion on ethics that is also ongoing by both academics and non-academics (Almeida et al., 2020; European Commission, 2020; Pagallo et al., 2019; Y. Wang et al., 2020; Winfield et al., 2019). Governments take an active role in monitoring AI and setting forth new regulations. Hurley (2018) focuses on how the US Department of Defense uses AI to better position itself for cyber events and challenges in the future. Their Responsible, Accountable, Consulted, and Informed (RACI) framework seeks to reconcile the imbalance between bringing in all relevant stakeholders and achieving organizational agility. Thus, while there is some work on stakeholders, a comprehensive view is missing. This follows, in particular, when investigating the governance cube in Figure 7, calling for structural, procedural, and relational mechanisms for stakeholder governance.

Furthermore, we also elaborated on the novel concept of governing "data value" on a more quantitative level, discussing costs, data quality, and model performance. These concepts also deserve more attention. Existing techniques provide a solid basis to assess the impact of data samples on model performance, which might serve as a surrogate for data value. But they lack predictive capability to answer questions such as "How much will our model (and data) be worth based on anticipated model performance?" Concept drift might render data worthless, but extant works primarily focus on reacting to concept drift rather than mitigating it proactively. Thus, it seems difficult to determine when data (or models) should be sold, updated, or acquired based on forecasts of their future performance. Even if data is not subject to concept drift, it is still likely to devalue over time, e.g., due to improved sensor capabilities such as cameras that capture higher resolution images that allow training ML models that perform better on specific object detection tasks.

## Conclusion

This study illuminated the building blocks of AI governance. Our work reviews existing literature on AI and ML using a conceptual framework. While most areas in AI governance deserve more attention in general, our review highlights areas that seem particularly interesting from a business perspective. Aside from a conceptual framework, we also provided more fine-granular views and proposals motivated by technological developments, e.g., related to data quality and value, and analogies to related disciplines, e.g., roles in software development and data governance. Future work can expand on these proposals, e.g., investigating governance roles and data value leveraging our AI governance cube providing a variety of options by picking a specific cell along the three dimensions "What is governed?," "How is it governed?" and "Who is governed?"

## Note

1. This can be computed from Alpha Go's paper (Silver et al., 2017) as done in https://www.yuzeh.com/data/agz-cost.html

## Disclosure statement

No potential conflict of interest was reported by the author(s).

## Notes on contributors

*Johannes Schneider* is an Assistant Professor in Data Science at the University of Liechtenstein. He has a Ph.D. in Computer Science from the Eidgenoessische Technische Hochschule (ETH) Zurich. Dr. Schneider's research has been published in journals such as European Journal of Information Systems (EJIS), Transactions on Software Engineering, Information Systems Management (ISM) and Data Mining and Knowledge Discovery(DMKD), and conference proceedings such as Foundations of Computer Science (FOCS), International Joint Conferences on Artificial Intelligence (IJCAI), Conference on Information and Knowledge Management (CIKM). His research interests encompass technical and managerial aspects of Artificial Intelligence and Data Science.

*Rene Abraham* is a Ph.D. candidate at the University of Liechtenstein. He studied Computer Science at Technische Universität Darmstadt and gathered work experience in the financial services industry. His research interest comprises data governance in intra-organizational and inter-organizational settings.

*Christian Meske* is a Full Professor of Socio-technical System Design and Artificial Intelligence at the Institute of Work Science and Faculty of Mechanical Engineering at Ruhr-Universität Bochum, Germany. His research on the design and management of digital innovations has been published in journals such as Business & Information Systems Engineering (BISE), Business Process Management Journal (BPMJ), Communications of the Association for Information Systems (CAIS), International Journal of Information Management (IJIM), Information Systems Frontiers (ISF), Information Systems Journal (ISJ), Information Systems Management (ISM), or MIS Quarterly Executive (MISQE).

*Jan vom Brocke* is the Hilti Endowed Chair of Business Process Management and Director of the Institute of Information Systems at the University of Liechtenstein. His work has been published in, among others, in Management Science, MIS Quarterly, Information Systems Research, Journal of Management Information Systems, Journal of the Association for Information Systems, European Journal of Information Systems, Information Systems Journal, Journal of Information Technology, and MIT Sloan Management Review.

## ORCID

Christian Meske http://orcid.org/0000-0001-5637-9433